\documentclass[a4paper,12pt]{article}
\usepackage[pdftex]{graphicx}
\usepackage{amsmath,amsthm, amsfonts, amssymb, amsxtra,url,natbib}
\usepackage[latin2]{inputenc}

\makeatletter
\renewcommand{\section}{\@startsection
{section}
{1}
{0mm}
{-3.5ex \@plus -1ex \@minus -.2ex}
{2.3ex \@plus .2ex}
{\rmfamily\raggedright\normalsize\bfseries}} %

\renewcommand{\subsection}{\@startsection
{subsection}
{2}
{0mm}
{-3.5ex \@plus -1ex \@minus -.2ex}
{2.3ex \@plus .2ex}
{\rmfamily\raggedright\normalsize\bfseries}} %
\usepackage{setspace}
\begin{document}
\def\figurename{}

\bibliographystyle{csplainnat}


\begin{center}
{\Large\textbf{AUGMENTED REALITY IMPLEMENTATION METHODS IN MAINSTREAM APPLICATIONS}}

\vspace{0.5cm}

David Procházka, Tomáš Koubek

 
\end{center}

\section*{Abstract} 

{\scshape PROCHÁZKA, D., KOUBEK, T.}: \textit{Augmented Reality Implementation Methods in Mainstream Applications}\\
Augmented reality has became an useful tool in many areas from space exploration to military applications. Although used theoretical principles are well known for almost a decade, the augmented reality is almost exclusively used in high budget solutions with a special hardware. However, in last few years we could see rising popularity of many projects focused on deployment of the augmented reality on different mobile devices. Our article is aimed on developers who consider development of an augmented reality application for the mainstream market. Such developers will be forced to keep the application price, therefore also the development price, at reasonable level. Usage of existing image processing software library could bring a significant cut-down of the development costs. In the theoretical part of the article is presented an overview of the augmented reality application structure. Further, an approach for selection appropriate library as well as the review of the existing software libraries focused in this area is described. The last part of the article outlines our implementation of key parts of the augmented reality application using the OpenCV library.

\vspace{0.5cm}
\noindent augmented reality, OpenCV, template matching, ARToolkit, computer vision, image processing. 

\section{Introduction}\label{Uvod}

The general principle of the augmented reality (AR) is embedding digital information into the real world scene. Thus it is a step between virtual reality and the real world. The embedded information is usually based on the content of the scene. A selected real object could be augmented by a virtual object or completely replaced. Well-known examples are the presentation of the fighter status report on the head-up display before a pilot or a navigation information projected on the wind shield of a car.

Applications based on the augmented reality have been used for decision making process support for many years. Military solutions for field operations made the pioneering work in this area -- from mentioned fighter head-up displays to tactical suits for troopers. Moreover there is a number of applications for construction and maintenance of complex systems. Especially space devices, planes and helicopters. A number of these solutions is outlined in \cite{ARManufacturing}. The discussed applications are usually developed for a single specific purpose. Especially from this reason, they are rather expensive. Therefore, they are used just for saving lives or speed-up production or maintenance of very expensive devices.

However, in last few years AR applications have been emerging into other areas -- design, medicine and even consumer electronics. Development of AR applications for this field is discussed in this article. It is obvious that one of the most important prerequisites for the success in this area is the price. This price is given by the price of the hardware (which is nowadays usually quite cheap) and the development costs. 

The development costs could be significantly cut-off using different image processing software libraries. AR applications are working on similar well-know principles. For a number of complex algorithms is therefore possible to use an existing implementation (image preprocessing, edge detection, etc.). The goal of this article is to present these libraries  potential for development of such mainstream AR applications and clearly outline the general AR application structure including the implementation in a selected library.

The section \ref{Hardware} shortly describes the augmented reality hardware and examples of its usage. In the section \ref{Metody} is outlined the key problem -- detection of an object in an image. The section \ref{SoftwaroveKnihovny} is a short review of the existing software libraries for image processing. In section \ref{Implementace} is presented our implementation of key parts of an augmented reality application.

\section{Hardware used for augmented reality solutions}\label{Hardware} 

Our reality could be augmented in many ways. Widely spread are for example audio navigation tools for visually impaired people. However, in the following review we will focus especially on a visual augmentation of the reality. This visual augmentation could be divided into three main categories. The first one is based on usage of the head mounted displays. The other group is based on projectors. This kind of augmented reality is called \textit{spatial augmented reality}. The last category is based on common displays (tablets, cell phones, etc.).

Currently used head mounted displays are based on the optical composition of the scene or on the video composition. The optical composition is a projection of artificial objects on a semi-transparent screen before user eyes. The video composition combines an image from a camera with artificial objects and the result presents on a small LCD screen in a virtual helmet. These semi-transparent screens are suitable especially for applications when a camera signal blackout could be critical (fighters, troopers, etc.). On the same principle there are in fact based head-up displays in cars with status and navigation information. The most important problem of this solution is the exact overlaying of a real and digital object. 

The solution is in the usage of the video composition based device. The image from the front-side camera is analysed, position of the real object is found and this object could be seamlessly replaced. Drawbacks of this solution are usually higher head mounted display size and weight, limited field-of-view and the price. Applications based on them are usually from the category discussed in the beginning of the article -- special high budget solutions. However, these solutions are not suitable for a common customer. 

The other group of products -- a solution based on projectors -- is significantly growing in last years. A computer is analysing the scene using a camera and searching for predefined objects. If the found, the attached projector is able to augment directly the real world object surface by the given digital information. A quite common example of this \textit{spatial augmented reality} is the adaptive projector used for projection on heterogeneous surfaces. The projected image is controlled by the software and adjusted to compensate the differences between the anticipated and the real image. A thorough description of this technology could be found in \cite{SAR}.

The last group of devices are solutions based on different screens. This area has been growing in the fastest way in last years. It is given especially by a huge emerge of advanced cellphones and tablets. These devices have all necessary components: a suitable display, a high resolution camera, a processor fast enough to make the real-time image analysis and also a GPS accompanied by compass. One of the most popular AR application of this kind is project \textit{Layar}\footnote{\url{http://www.layar.com/}}. From the technical point-of-view, it is a video composition based application merging the camera image with additional information from map layers stored inside the device. 

The cellphones and other portable devices present a platform with a significant economical potential. The number of users is in comparison to previously mentioned categories incomparable. According to the research done by the Garther agency, 62 million \textit{smartphones} with ability to run such AR applications was sold only in second quarter of the year 2010. The rapidly growing market of tablets founded by the Apple's \textit{iPad}\footnote{\url{http://www.apple.com/ipad/}} is also important.

\subsection{Comparion of AR implementations}\label{RealizaceAR}

Although the presented output devices are completely different, there is a number of common principles. The first key problem is the identification of the screen before the user. For this purpose could be used solely image processing (searching for known objects) or there could be used other position techniques. There is frequently used the triangulation from a cellphone network, Wi-Fi hotspots, the GPS or some inertial sensors. At the moment the scene before the user is identified, it is necessary just to insert appropriate information. These first two steps are based on well-known common principles described later. 

A significant difference is in the method of presentation to the user. However it is just a question of the used hardware. Software architecture of the systems is usually very similar.

\section{Object detection methods}\label{Metody} 
\begin{figure}[h!t]
	\centering\includegraphics[width=150mm]{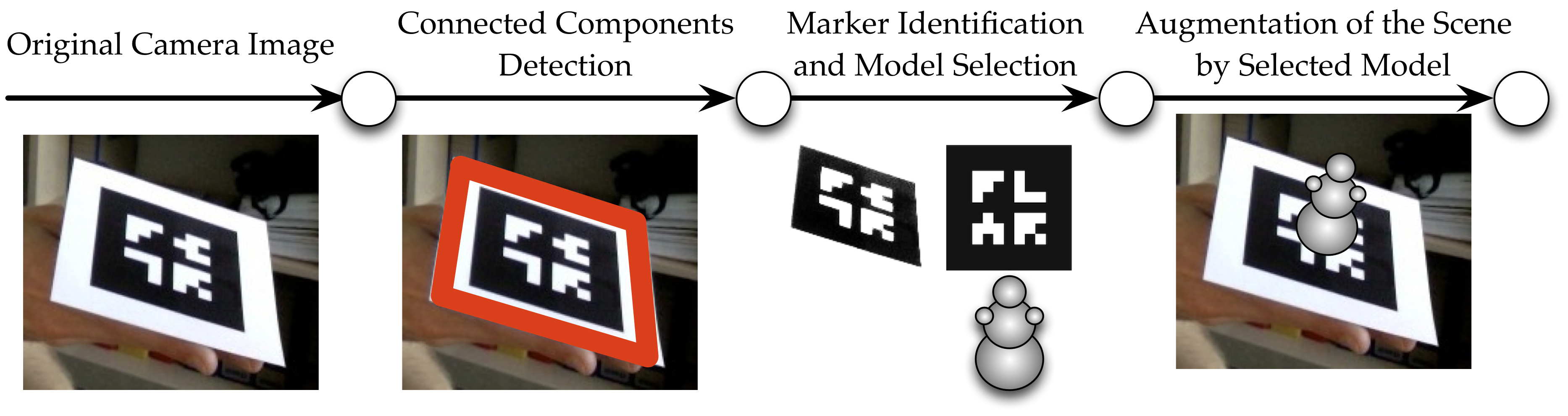}
	\caption{Scheme outlining the basic functionality of the augmented reality application.}
	\label{obr:zakladni}
\end{figure}

The general structure of any application based on image processing is following: We will  acquire an image from a camera and store it into an inner representation (a kind of RGB color matrix). Further we will make an image analysis and identify a possible wanted object, its position and orientation. This potentially wanted object is compared with a predefined pattern or patterns. In the case of success, the last step is insertion of the artificial object. This process is illustrated on the fig. \ref{obr:zakladni}. This process is quite common for all AR applications.

Another significant difference is especially in the step of comparison of the possible desired object with the patterns. It depends whether searched object is a face, a natural object (building), a simple shape (window) or an artificial marker (usually black and white square with a predefined pattern). In the following part of the article we will focus on the artificial marker. This case is quite simple and could be the first step in building of a robust AR application.

\subsection{Artificial marker detection}\label{DetekceMarkeru} 
\begin{figure}[h!t]
	\centering\includegraphics[width=120mm]{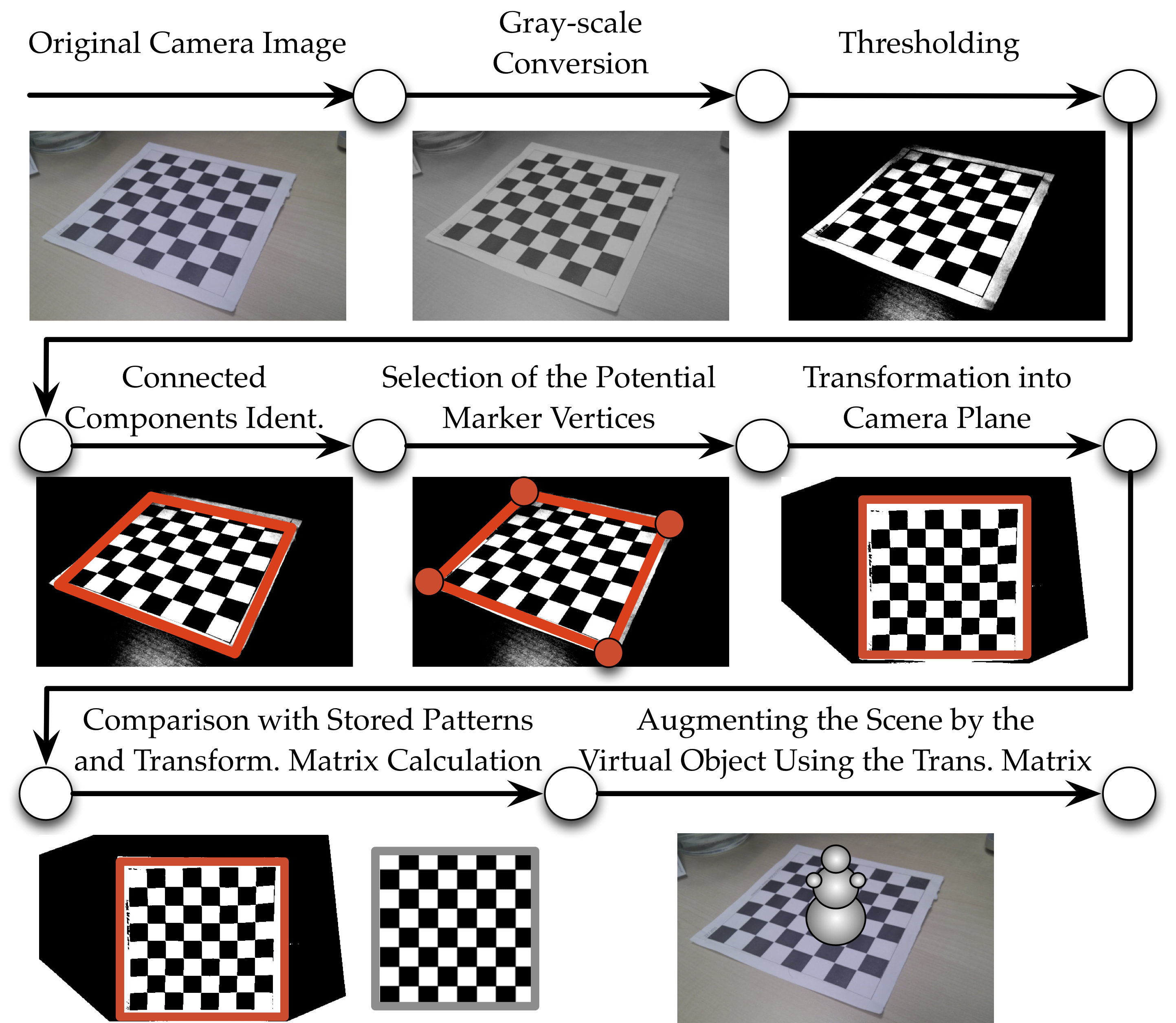}
	\caption{Stages of object detection and augmenting of the scene.}
	\label{obr:podrobne}
\end{figure}

An example of the marker detection process could be described by following steps. The whole process is also outlined on Fig. \ref{obr:podrobne}. Our presented method is not the only possible solution, however it is widely used by many well tested applications.

As been already mentioned, from an input device a color image is taken and stored into the inner representation of the image processing library. Further, this image is transformed into the gray scale. It is possible to make a standard conversion of all color channels or prefer a specified color channel (e.g. green). The grey value now presents the brightness of the pixel (hence an object).

Such image contains a number of objects -- markers, persons, furniture, etc. For  performance improvement it is necessary to remove most of these objects from the image. This step is usually done via thresholding. Simple thresholding could be generally described using following formula:

\begin{equation}
g(i,j) = \left\{
\begin{matrix}
0, \;\; f(i,j)\leq P\\ 
1, \;\;  f(i,j)> P 
\end{matrix}\right.
\label{eq:prahovani}
\end{equation}
where function $f(i,j)$ is the source image (brightness of the pixel), value $P$ is the threshold and $g(i,j)$ is the result image. Value of the threshold is usually determined according to the scene content. 

The gist of the thresholding is transformation to a bitmap in such way that allows to remove most of unnecessary objects. A well chosen thresholding method could significantly improve the performance of the application. Each object that is not filtered out is a potential marker and therefore must be tested as described further. If its possible, it is recommended to prepare also an appropriate testing environment. Suitable lighting and high contrast markers could significantly simplify the preprocessing phase. Generally, a homogeneous controlled environment allows to wipe out most of the unwanted objects (se e.g. \cite{homo}). In case of complex lighting conditions, different adaptive thresholding techniques are usually used.

The result of the previous step is an image with number of vertices and edges. The other step is detection of connected components with the required shape. This could be done using an image morphology algorithm. This algorithm produces a list or tree of image components. For identification of potential markers it is sufficient to browse this list of objects and test whether the provided entity fulfils the given criterion. In case of a common white square marker with a black inner square with a pattern, it is an entity with four vertices with an inner entity again with four vertices.

As soon as we have potential marker vertices, they must be compared with predefined patterns. For this comparison there must be done marker pixels transformation between a marker plane and a camera plane (in other words: perspective distortion must be eliminated). Equation \ref{eq:transformace} describes this transformation. If defined transformation matrix is applied on point [$x_{m}$, $y_{m}$, $z_{m}$] in marker plane, we will receive the position of this vertex in the camera plane. By inversion of this process we could receive the original vertex position before perspective distortion (generally we will receive the original object shape). Elements $T_1$ -- $T_3$ represent a translation vector. Elements $R_{11}$ -- $R_{33}$ represent well known 3$\times$3 rotation matrix (see \cite{OpenGL}, p. 806). Calculation of the transformation matrix elements is described in \cite{actaVRUT}. By this step we have restored the original shape of the object and it is possible to make comparison with the marker patterns.

\begin{equation}
\left[
\begin{array}{c} 
x_{c} \\ 
y_{c} \\
z_{c} \\
1 \\
\end{array}
\right]=
\left[
\begin{array}{cccc} 
R_{11} & R_{12}   & R_{13}   & T_{1} \\ 
R_{21} & R_{22}   & R_{23}   & T_{2} \\ 
R_{31} & R_{32}   & R_{33}   & T_{3} \\ 
0	   & 0   	  & 0		 & 1 \\ 
\end{array}
\right]
\left[
\begin{array}{c} 
x_{m} \\ 
y_{m} \\
z_{m} \\
1 \\
\end{array}
\right]
\label{eq:transformace}
\end{equation}

A precondition of a successful transformation is a camera calibration step. The camera calibration matrix describes optical properties of a given device and compensates also possible optical errors. This calibration matrix is calculated for a given camera only once. Therefore, the performance of the application is not affected. The application of calibration matrix is described in \ref{eq:kalibrace}. Its structure and calculation is outlined in \cite{Kato99}.

\begin{equation}
\left[
\begin{array}{c} 
x_{c} \\ 
y_{c} \\
z_{c} \\
1 \\
\end{array}
\right]=
\left[
\begin{array}{cccc} 
C_{11} & C_{12}   & C_{13}   & C_{14} \\ 
C_{21} & C_{22}   & C_{23}   & C_{24} \\ 
C_{31} & C_{32}   & C_{33}   & 1 \\ 
0	   & 0   	  & 0		 & 1 \\ 
\end{array}
\right]
\left[
\begin{array}{cccc} 
R_{11} & R_{12}   & R_{13}   & T_{1} \\ 
R_{21} & R_{22}   & R_{23}   & T_{2} \\ 
R_{31} & R_{32}   & R_{33}   & T_{3} \\ 
0	   & 0   	  & 0		 & 1 \\ 
\end{array}
\right]
\left[
\begin{array}{c} 
x_{m} \\ 
y_{m} \\
z_{m} \\
1 \\
\end{array}
\right]
\label{eq:kalibrace}
\end{equation}

The last step -- marker identification -- is the most time consuming operation. It is a correlation calculation between the potential marker image and the pattern. In case matching is above the given limit, images are taken as corresponding. For the correlation calculation could be used a number of methods. From neural networks to least squares algorithm. Selected approaches are described in \cite{Kato03}. Methods implemented directly in \textit{OpenCV} library could be found in \cite{OpenCV} on page 214. Each potential marker must be tested against all patterns using a selected method until all patterns are tested or the appropriate pattern is found. It is obvious that computation complexity grows linearly with the number of patterns. That is the reason why in many applications there are used special markers with patterns given by algorithms such as the well known Golay error code.

In case there is a corresponding pattern, the application will store the transformation matrix that defines the orientation of the marker in the scene. This matrix is the homomorphy matrix described before.

\subsection{Conclusion of marker identification process}
Basic principles of the AR applications are quite similar, as is obvious from the  description outlined above. The difference is mostly in particular algorithms for image comparison. That is the reason why it is not effective to implement the application from the scratch, but to use an existing library that supports the mentioned well-known algorithms. The following part of the article describes an appropriate library selection.

\section{Methodics of image processing library selection}\label{SoftwaroveKnihovny} 
Before library selection for our AR application it is necessary to specify exactly the functionality which is required. For a huge number of applications, following criteria are important:

\begin{enumerate}
\item Required programming language support: Fulfilment of this criterion could be complicated. Most of the libraries support just C/C++. An exception is the \textit{OpenCV} library that supports also the Python language and \textit{NyARToolKit} supporting many mainstream languages. 

\item Required platform and architecture support: This problem is obviously the most limiting criterion. Especially in case application is targeted on mobile devices. On personal computers there is limited only support of 64 bit architecture that is necessary for complex applications.

\item Project is under active development: There is a huge number of already unsupported projects or projects with very limited numbers of active developers and users. Usage of such library is not recommended. Support of new architectures, input devices, etc. is an essential feature of any library. 

\item Documentation: This aspect is in the beginning of the projects frequently underestimated. However, developers will be facing a number of situations where is utmost important to understand thoroughly the implemented method. It is not enough to know that this method or function if proving some correlation coefficient, it is necessary to know exactly how the algorithm works.

\item Provided functions number: Again, it is quite an obvious requirement. It is recommended to consider the application future development. To base the project on a library with limited functionality could be in long term point-of-view expensive.
\end{enumerate}
Further section briefly reviews selected frequently used image processing toolkits.

\subsection{OpenCV}\label{OpenCV}

\textit{OpenCV}\footnote{More information on: \url{http://opencv.willowgarage.com}} (\textit{Open Computer Vision}) is an open source toolkit for the real-time image processing. \textit{OpenCV} is the most robust solution among all frameworks in comparison. It supports C++, C and Python languages. It consists of tools for image analysis, image transformations, camera calibration, stereo vision and also tools for simple graphical user interface and more.

\textit{OpenCV} is associated with \textit{Intel} company, which implements the support for \textit{OpenCV} into hardware. This library uses \textit{Intel Integrated Performance Primitives} that provides high performance for low-level routines for sound, video, speech recognition, coding, decoding, cryptography etc. \textit{Intel Threading Building Blocks} is used for parallel processing.

The library is a cross-platform, there are versions for GNU/Linux, Microsoft Windows and Mac OS X, 32b and 64b systems. A big advantage of this project is that it is still in development. In December 2010 there was released version 2.2. The development of a new version is in progress. The community of \textit{OpenCV} users is huge, there are a lot of manuals, tutorials and discussion forums. Basics of \textit{OpenCV}, mathematical principles of image processing and their implementation are described by \cite{OpenCV}.

Easy usage of cameras is very useful for augmented reality applications development. This framework uses resources of operating systems and communication with hardware drivers is fully provided as well. If a camera driver is installed in an operating system, it can be initialized and used immediately. Disadvantage of this library is that there are not implemented any direct methods for marker registration. As shown below, for this purpose it is necessary to link several different \textit{OpenCV} functions.

\subsection{ARToolKit}\label{ARToolKit}

\textit{ARToolKit}\footnote{\url{http://www.hitl.washington.edu/artoolkit}} is a software library for development of augmented reality applications. It is a cross-platform. There are versions for GNU/Linux, Microsoft Windows, Mac OS X and SGI, but officially only for 32b versions of the systems. This framework implements some basic tools for marker registration. This is more straightforward in comparison with the \textit{OpenCV}. Implemented methods are robust. The library supports C language. The community of \textit{ARToolKit} users is not as big as \textit{OpenCV} community, but still quite large.

Working with cameras is more complicated than in the case of \textit{OpenCV}. However, a more substantial problem is that development of a free version of this framework was stopped and it continues only for the paid version called \textit{ARToolKit Professional}\footnote{\url{http://www.artoolworks.com}}. The last free version was released in February 2007 and since then no further update has been released. The development is stopped and probably no new version will be released in the future. Compatibility with new versions of the operating systems or other features is not guaranteed. These can be found in the paid version, which is very costly. The development license costs 4995\$ a year, use of the framework in third party applications costs 995\$ a year. The development of a commercial version is currently uncertain, because the last major changes of \textit{ARToolKit Professional} were made, according to the documentation, in 2007. The library is still maintained.

\subsection{NyARToolKit and derivates}

\textit{NyARToolKit}\footnote{\url{http://nyatla.jp/nyartoolkit}} was derived from \textit{ARToolKit} version release 2.72.1 (the last version of free \textit{ARToolKit}). Nowadays is \textit{NyARToolKit} developed by Japanese author. In fact, it is derivated from \textit{ARToolKit} for other platforms and languages (\textit{ARToolKit} is for C language only). There is a version for Java, Flash, Android, Silverlight (\textit{SLARToolKit}), Actionscript (\textit{FLARToolKit}), or C\#. C++ version is in beta stage (no support for 3D or cameras). There are still some restrictions, i.e. documentation is in Japanese but project is active and other shortages can be removed in next versions. The community will be crucial.

\subsection{Other AR libraries}

Other frameworks for working with the AR are \textit{Morgan}, \textit{DART}, \textit{Goblin XNA} or \textit{Studierstube}. Mostly these are university projects, but they are not so recent or their support is only marginal, as well as their user community is only limited. \textit{Studierstube} is worth of noticing, because of the best documentation (among university projects). It is developed by Institute for Computer Graphics and Vision at Graz University of Technology. At first \textit{Studierstube} has been used for collaborative AR, later was focused on mobile applications. This library uses other frameworks for tracking, video and registration.

It is advisable to check the development of \textit{Goblin XNA}. It is a Columbian University project, derived from the project \textit{Goblin} with added support for Microsoft XNA framework. \textit{ARTag} is used for marker tracking. Connection with XNA helps to bring augmented reality applications on the Microsoft Xbox platform. However, it is also available for GNU/Linux and MacOS X.

\subsection{Choice of appropriate library}

\textit{OpenCV}, \textit{ARToolKit (Profesional)}, \textit{NyARToolKit} and \textit{Studierstube} have some pros and cons, but they belong to better ones. But none of these fulfilled completely specified criteria for the framework choice. All frameworks are cross-platform. Support for 64b is provided for \textit{OpenCV}, \textit{NyARToolKit} and \textit{ARToolKit Professional}. \textit{OpenCV}, \textit{ARToolKit (Professinal)} and \textit{Studierstube} have a lot of documentation (\textit{OpenCV} has both official documentation and literature).

\textit{OpenCV} has the best stability of development among all frameworks. New versions are released often and a lot of mistakes are corrected and a new features are added. For other projects, continuation of the development is uncertain. \textit{NyARToolKit} publishes updates approximately in six-month intervals, the last in April 2011. \textit{Studierstube} released the latest version of the framework two years ago.

The choice of a library is not clear at all, because it depends on particular needs of the project. However, for production deployment \textit{NyARToolkit}, \textit{ARToolKit} and \textit{Studierstube} are not appropriate. The big disadvantage is the uncertain or stopped development. These libraries have big potential, but there should be some issues in future, especially new architecture or operating system compatibility etc. The development of the \textit{ARToolKit Professional} is also questionable. The project website was updated at 2011, but only version 2.72.1 (from 2007) is avalaible for free.

The choice of an appropriate library is a very import decision at this point. From our point-of-view, there are two options -- \textit{ARToolKit Professional} and \textit{OpenCV}. The first one is ready for development augmented reality applications, but it is very expensive. \textit{OpenCV} has a good documentation, user support and it is free, but not ready for AR applications development. The user has to implement a method for marker tracking or recognition. However, the \textit{OpenCV} providea a lot of image processing functions which can be useful later. We decided to use the \textit{OpenCV} for development of our AR application.

\section{Detection of artificial marker in OpenCV environment}\label{Implementace} 

As we described earlier, the \textit{OpenCV} library does not implement any direct methods for identifying and registering artificial markers in space. But it provides a lot of methods for image analysing and image processing, which can be used for implementation of marker recognition and registration. For testing of the AR applications \textit{OpenCV} offers functionality for finding of the special type of marker -- chessboard. The use of chessboard marker makes development of simple AR applications much easier. Next section describes main ideas for finding of a artificial marker. The method details can be found in \cite{OpenCV}, in \textit{OpenCV} reference manual on the project homepage and in code examples distributed within the installation package.

\subsection{Finding of artificial marker vertices}\label{HledaniVrcholu}

As mentioned in section \ref{Metody}, firstly the image has to be converted to gray-scale and thresholded. Gray-scaling is done by function \texttt{cvCvtColor} which converts \textit{OpenCV} internal format to different color spaces. This conversion is possible to color spaces RGB, CIE Luv, CIE Lab, HLS, HSV, YCrCb and CIE XYZ. One parameter of \texttt{cvCvtColor} specifies between which two color spaces the image is converted. To threshold image function \texttt{cvAdaptiveThreshold} can be used. Its output is a binary image. The threshold value, threshold methods and their settings are specified in the function parameters.

Next step of our analysis is finding of edges and vertices. \textit{OpenCV} provides several functions which implement different algorithms for these actions. For edge detection is available i.\,e. Canny edge detector or Hough transformation. Mentioned methods are described at \cite{1682677}, respectively \cite{361242}. Another option is \textit{OpenCV} function \texttt{cvFindCountour}. The process described below works with well-known artificial marker -- a black rectangle within a white field, with a unique picture (can be compared with template).

Our application uses the \texttt{cvFindContour} function to find contours in the thresholded image. There are two kinds of contours -- an inner and outer contour. All contours are represented by an \textit{OpenCV} structure called \texttt{cvSeq} and is stored into the special storage. Some parameters of \texttt{cvFindContour} are important for the whole algorithm. Parameter \texttt{mode}, in function \texttt{cvFindContour}, means, how the found contours will be organized. If \texttt{mode} is set to \texttt{CV\_RETR\_CCOMP}, all contours are organized into two level hierarchy -- parents and children (it is possible to obtain also a list or tree of contours). 

All contours are stored in the internal storage, and we must decide, which contours belong to a particular marker. At first, we can approximate polygon from contours by function \texttt{ApproxPoly}. It is an important step in finding a marker, bacause polygons have some additional attributes, i.\,e. attribute \texttt{total}, that expresses amount of lines of a polygon. 

Choice of right contours is crucial. We can use attributes of \texttt{cvSeq} and attributes of polygons. We declare, that a contour is a marker, if it consists of 4 lines (being the attribute of polygon -- determining outer contours of the black rectangle), if has a child (being the attribute of \texttt{cvSeq} -- determining inner contours of black rectangle) and the child consists of 4 lines. These conditions determine the edges of the marker. Each contour is the input for structure \texttt{CvSeqReader}. This is used for finding vertices of markers. For this purpose there is called macro \texttt{CV\_READ\_SEQ\_ELEM}. This algorithm output is a marker vertices quaternion. It is used for transformation of the coordinate system (see \ref{obr:podrobne}, step 6). Now we have a quadruple of points that describe the vertices of a potential marker.

\subsection{Transformation of vertices coordinates into camera plane and image matching}

For matching of the found object with an image template it is necessary to compensate its perspective transformation. For computing of the transformation matrix, which describes this marker pose and position, function \texttt{cvFindHomography} can be used. Its input is the matrix of points which represents the marker template and matrix of points which represents the marker in the image (this is just a square of given size). Their transformation is described by the matrix in equation \ref{eq:transformace}. It is also used for compensation of marker rotation. For this \texttt{cvWarpPerspective} is used. This function makes inverse operation for transformation which is described by the matrix (for inverse transformation is necessary to use flag \texttt{CV\_WARP\_INVERSE\_MAP}). The output of this function is an image without perspective transformation. This image can be matched with the template.

For comparison of two images \textit{OpenCV} provides own implementation of a template matching method, described by \cite{match}. It is implemented in function \texttt{matchTemplate}. The template slides through the image, compares the overlapped patches against the template using a specified method and stores the comparison results to a matrix (result value expresses probability of the template presence in the image and depends on the matching method). For finding the template position in the image function \texttt{minMaxLoc} must be called. This function finds the position and value of global minimum and maximum in the result matrix.

Surely the match \texttt{matchTemplate} method is not the only solution for pattern recognition. There could be used a number of methods including different clustering methods (see e. g. \cite{fejfar}) or neural networks (image processing applications outlined e. g. in \cite{mendel2011}). 

\subsection{Finding chessboard marker}

\begin{figure}[h!t]
	\centering\includegraphics[width=140mm]{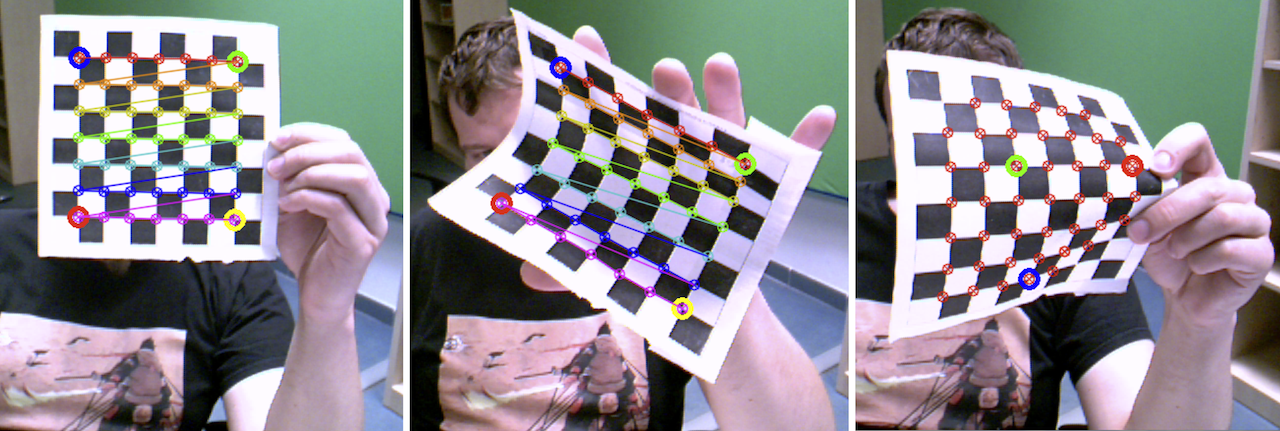}
	\caption{Example of chessboard marker detection.}
	\label{obr:rohy}
\end{figure}

In this section there is implemented detection of chessboard vertices. This is a special case of the issue described in section \ref{HledaniVrcholu}. For this chessboard vertices detection \textit{OpenCV} includes function \texttt{cvFindChessboardCorners} which finds all inner corners of the chessboard. The input of this algorithm is the chessboard size -- amount of the inner corners (height and width of the chessboard). There are few methods for finding the corners. Function \texttt{cvFindCornerSubPix} can be used for a more precise determination of the corner position. After this, the position of all inner corners of the chessboard is stored into an array of \textit{cvPoint}. Vertices, which represent the four edge vertices of the marker are on these array coordinates: \texttt{0},\texttt{ boardWidth-1}, \texttt{(boardHeight-1)*boardWidth} and \texttt{(boardHeight-1)*boardWidth+boardWidth-1}, number \texttt{boardWidth} represents the amount of inner chessboard corners in rows and \texttt{boardHeight} represents the amount of corners in columns. This algorithm output is a marker vertices quaternion. These are the corners coordinates of the chessboard marker in the image. These corners are highlighted in fig. \ref{obr:rohy} by big circles. Other found corners are highlighted by smaller circles. There is an apparent robustness for rotation and bending of the chessboard. A loss of inner corners occurs at big deformation of the marker.

\section*{Discussion and conclusions}

The implementation clearly shows that the \textit{OpenCV} library is able to implement same functionality as the well-know \textit{ARToolkit} and other libraries based on this project. This is a key issue for many developers considering the AR application development. According to our experience, the \textit{OpenCV} is more feasible solution than the \textit{ARToolkit}, despite more complicated beginnings.

The augmented reality applications for common users are emerging area. Low user-friendliness is a crucial problem of the development, although there is a significant research in this area for almost two decades. For instance, there are no standard approaches for user interface design, even despite the fact these are commonly used for a desktop and mobile applications design -- see \cite{design}, \cite{krystof1} and many others. One of the reasons is that many metrics and design patterns are not applicable to AR applications development. Assimilation of new metrics and patterns is a significant challenge. In future work we want to focus on this area. 

\section*{Summary}

As been outlined in our article, all augmented reality applications work on similar theoretical principles. The crucial difference is in a position/orientation detection and in a template matching approach. 

Only satellite navigation systems, compasses and motion sensors are used in a part of applications. However, these applications usually do not allow representation of complex graphics objects. The other group of applications use the image analysis to specify the information about orientation of the user. The base of these applications lies in composition of a signal from a camera and a digital information. In our article we deal with this area. We chose the problem of an identification of an artificial marker for illustration. 

The functionality of the application based on the image analysis was defined in the section \ref{DetekceMarkeru} as the ability of: image reading from an input device, its preprocessing (gray scale transformation, thresholding), segmentation on continuous objects, vertex detection, compensation of geometric distortions and the comparison with given patterns. The list of these processes could be supplemented by an ability to insert the model/information to the image. Its implementation depends on the architecture of the application (if OpenGL, Microsoft DirectX or other API is used). The aim of this section is to explain comprehensibly basic principles of the image processing in AR applications.

The basic implementation scheme, which is briefly described in section \ref{Implementace}, presents clearly the implementation of discussed problems using the \textit{OpenCV} library. The usage of presented solutions can reduce development time. Discussed projects (and especially the \textit{OpenCV} project) are open-source.
Therefore, it is possible to modify or extend the unsatisfactory existing implementations of the methods. On the basis of our experience with discussed libraries, we recommend to use the \textit{OpenCV} library.  The significance of the project, open source codes, high-quality implementation, available documentation and also the wide community and support of Intel company are indicators of project quality and to some measure guarantee of further development.

\section*{Acknowledgements} 

This paper is written as a part of a solution of project IGA FBE MENDELU 31/2011 and research plan FBE MENDELU: MSM 6215648904.

\bibliography{biblio}

\section*{Address}
Ing. David Procházka, Ph.D., Department of Informatics, Mendel University in Brno, 61300 Brno, Czech Republic, email: david.prochazka@mendelu.cz

\end{document}